\documentclass[11pt,letterpaper]{article}
\usepackage{emnlp2017}
\usepackage{times}
\usepackage{latexsym}
\usepackage{amsmath}
\usepackage{float}
\usepackage{graphicx}
\usepackage{url}
\emnlpfinalcopy

\def\emnlppaperid{***}

\title{GradAscent at EmoInt-2017: Character- and Word-Level Recurrent Neural Network Models for Tweet Emotion Intensity Detection}

\author{Egor Lakomkin\Thanks{equal contribution}, Chandrakant Bothe\footnotemark[1] \and Stefan Wermter \\
Knowledge Technology, Department of Informatics,\\
University of Hamburg,\\
Vogt-Koelln Str. 30, 22527 Hamburg, Germany\\
\url{knowledge-technology.info} \\
{\tt \{lakomkin, bothe, wermter\}@informatik.uni-hamburg.de}}

\date{}

\begin{document}

\maketitle

\begin{abstract}
The WASSA 2017 EmoInt shared task has the goal to predict emotion intensity values of tweet messages. Given the text of a tweet and its emotion category (anger, joy, fear, and sadness), the participants were asked to build a system that assigns emotion intensity values. Emotion intensity estimation is a challenging problem given the short length of the tweets, the noisy structure of the text and the lack of annotated data. To solve this problem, we developed an ensemble of two neural models, processing input on the character. and word-level with a lexicon-driven system.  The correlation scores across all four emotions are averaged to determine the bottom-line competition metric, and our system ranks place forth in full intensity range and third in 0.5-1 range of intensity among 23 systems at the time of writing (June 2017).
  
\end{abstract}

\section{Introduction}

\par Sentiment analysis of a text reveals information on the degree of positiveness or negativeness of the opinion expressed by the writer. Such information can be useful for providing better services for users \cite{kang2014review} or preventing potentially dangerous situations \cite{o2015detecting}. Traditionally the most popular way of sentiment representation is either binary (positive, negative) or multi-class (for example 5 classes: very negative, negative, neutral, positive, very positive). While being simple, such a scheme looses interpretability and a continuous intensity scale might be preferred. Twitter sentiment and emotion intensity detection are still challenging tasks and remain active areas of research. These difficulties have several reasons: extensive usage of hash-tags, slang, abbreviations, and emoticons. Also, tweets are usually typed on mobile devices which can lead to a substantial amount of typos. As traditional NLP tools are usually trained on datasets containing clean text, which makes it difficult to use them for tweet analysis.  

\par Existing approaches for modeling emotion intensity rely heavily on manually constructed lexicons, which contain information about intensity weights for each available word \cite{mohammad2017emotion,neviarouskaya2007textual}. The intensity for the whole sentence can be inferred by combining individual scores of words. While being easily interpretable, such models have several limitations. Ignoring word order and compositionality of the language is the first issue, which is critical for modeling sequences. Constructing such lexicons is a labour-intensive process, which needs to be carried out continuously due to the constant development of language. Data-driven approaches like deep neural networks can overcome such limitations, and they have been behind many recent advances in text processing tasks, such as language modeling, machine translation, POS tagging, and classification \cite{Irsoy2014OpinionNetworks,Socher2013RecursiveTreebank}. The appealing property of such models is their ability to combine feature extraction and classification stages given a sufficient amount of training data.

\begin{figure*}[!t]
\centering
\includegraphics[height=5.8cm, width = 14.5cm]{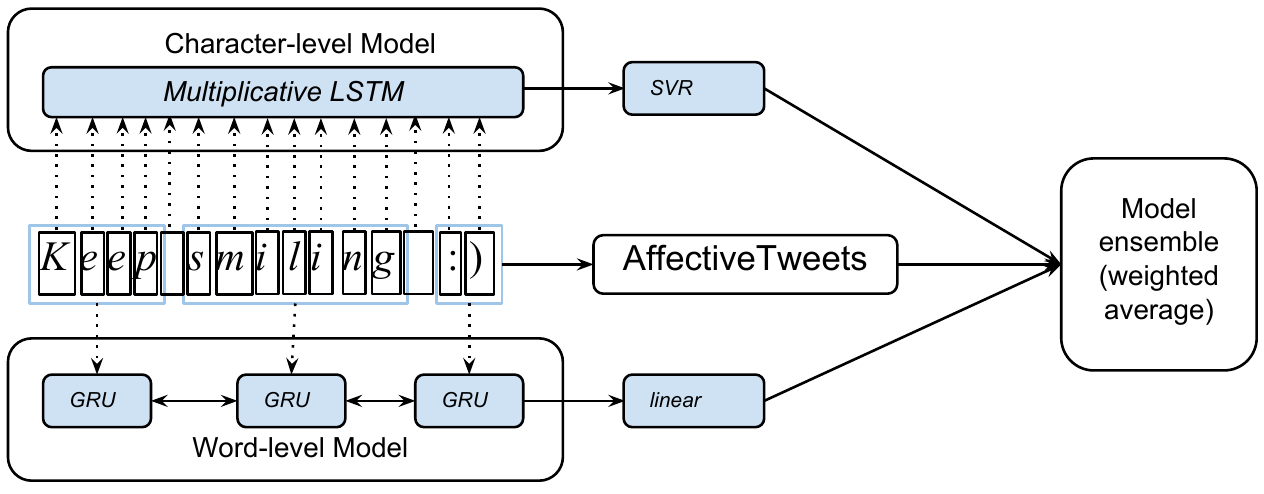}
\caption{Overall model architecture. It combines a lexicon-based AffectiveTweets model with two neural models: a character and a word-level model via averaging scores with weights tuned on the provided validation set.}
\label{fig:rnnlstm}
\end{figure*}

\par In this paper, we augment traditional lexicon-based models with two neural network-based models: one with character and one with word input. Character-level deep neural networks recently showed outstanding results on text understanding tasks such as machine translation \cite{kalchbrenner2016neural} and text classification \cite{Zhang2015Character-levelClassification}. In a domain-specific task such as predicting the emotion intensity of tweets, a character-level model can theoretically capture the notion of hashtags, emoticons, or character repetitions, which all are unique to social media. The intuition is that a character-level model captures common writing patterns such as punctuations and signaling characters. A word-level recurrent neural model can incorporate the order of information using distributed representations of words trained on a large amount of text. 

\par Our final model is a weighted average of the scores provided by the baseline, our character- and word-level model. Our ensemble model achieved forth position in the 0-1 emotion intensity range task and third position in the 0.5-1.0 range task on the public leaderboard (GradAscent team) on CodaLab\footnote{https://competitions.codalab.org/competitions/16380} at the time of writing this paper (June 2017).

\section{Approach}

Our system is an ensemble of the provided baseline system and two neural network-based models; processing character and word input respectively. Combining the word and character representations we can deal with noisiness of the tweet messages as well as capturing the semantics of the text by using distributed word representations. 

\subsection{Data pre-processing}

We perform only a few preprocessing steps, like striping URLs, user mentions (@username) and leave only the following characters: \verb|a-zA-Z@-!:(),;?.#'0-9*|. We always convert a message to lowercase before feeding it to the models.

\begin{table}[!htbp]
\centering
\caption{WASSA 2017 Emotion Intensity Shared task dataset statistics. }
\label{dataset-statistics}
\bigskip
\begin{tabular}{l*{5}{c}}

\textbf{Split}              & \textbf{Joy} & \textbf{Anger}  & \textbf{Fear}& \textbf{Sadness} & \textbf{Sum} \\
\hline
\textbf{Train}       & 823   & 856  & 1147 & 786  & 3612 \\
\textbf{Dev}         & 78    & 83   & 109  & 73   & 343  \\
\textbf{Test}        & 714   & 760  & 995  & 673  & 3142 \\

\hline
\end{tabular}
\end{table} 

\subsection{Baseline model}

The baseline system is a WEKA-based model called AffectiveTweets \cite{mohammad2017emotion}. This system combines features derived from several lexicons like MPQA \cite{DBLP:conf/naacl/WilsonWH05}, Bing Liu \cite{DBLP:conf/kdd/HuL04}, AFINN \cite{DBLP:conf/msm/Nielsen11}, Sentiment 140 \cite{NRCJAIR14}, NRC Hashtag sentiment lexicon, NRC Word-Emotion Association Lexicon \cite{Mohammad13}, NRC-10 Expanded \cite{DBLP:conf/webi/Bravo-MarquezFM16}, NRC Hashtag Emotion Association \cite{COIN:COIN12024}, and SentiWordNet \cite{DBLP:conf/lrec/BaccianellaES10} with traditional NLP features like word- and character n-grams, POS tags \cite{DBLP:conf/acl/GimpelSODMEHYFS11}, and processing of negations. In addition to those features, AffectiveTweets incorporates SentiStrength values \cite{DBLP:journals/jasis/ThelwallBP12}, Brown clusters \cite{brown1992class} trained on $\sim$53 million tweets\footnote{\url{http://www.cs.cmu.edu/~ark/TweetNLP/}}, combining them with averaged and concatenated first $k$ word embeddings of the tweet. Finally, a support Vector Machine model is used as a regression model for predicting emotion intensity values.

\subsection{Character-level RNN model}

\par We extracted character-level sentence representations by encoding the whole tweet text with the pre-trained recurrent neural network model\footnote{\url{https://github.com/openai/generating-reviews-discovering-sentiment}}. This model contains a single multiplicative LSTM \cite{krause2016multiplicative} layer with 4,096 hidden units, trained on $\sim$80 million Amazon product reviews as a character-based language model \cite{Radford2017LearningSentiment}. We extracted the hidden vector corresponding to the last character of a tweet and also averaged the representations of all hidden vectors. Concatenation of the two vectors is used as a tweet representation. In our experiments, we observed that adding averaged character representations improves the overall performance, especially when evaluating high-intensity tweets.

\par In addition to the pre-trained character-level language model, we investigate a model trained specifically for tweets. Our observation was that the tweets have a different language structure than product reviews, which might affect the transferability of features between domains. 
For instance, the extensive use of emoticons, character repetition, and hashtags, which are common for tweet messages, however, significantly different from product reviews which are often longer and grammatically correct. 

We trained the character-based language model on the Sentiment 140 corpus comprised of 1.6 million tweets \cite{go2009twitter}. A single-layer LSTM \cite{hochreiter1997long} with 1024 hidden units was trained with Adam optimizer \cite{kingma2014adam} with 0.0005 learning rate and clipping gradients at norm 1. 
We used the Support Vector Regressor (SVR) algorithm to classify tweets represented as a fixed-length vector with a character-based recurrent neural network. Results of different setups are reported in Table \ref{char-lm}.

\begin{table}[!t]
\centering
\caption{Effect of different character-level recurrent neural network representations: last cell vector of the pre-trained model (PT, last) and Twitter-specific character LM (Twit, last). Also, in addition, we tested a concatenation of the last cell vector with the average of all cell vectors for the pre-trained model (PT, last+avg) and Twitter model (Twit, last+avg). Results are reported on the test set, where avg\_p corresponds to Pearson coefficient and avg\_s to Spearman. }
\label{char-lm}
\bigskip
\begin{tabular}{l*{4}{c}}
\textbf{Range}        & \multicolumn{2}{c}{\textbf{(0.0-1.0)}}   &  \multicolumn{2}{c}{\textbf{(0.5-1.0)}} \\
\textbf{Model}  & \textbf{avg\_p} & \textbf{avg\_s} & \textbf{avg\_p} & \textbf{avg\_s} \\
\hline
\textbf{PT, last}        & 0.470 & 0.468 & 0.412 & 0.404 \\
\textbf{PT, last+avg}    & \textbf{0.474}   & \textbf{0.472}   & \textbf{0.419}   & \textbf{0.413}   \\
\textbf{Twit, last}      & 0.312   & 0.307   & 0.296   & 0.288   \\
\textbf{Twit, last+avg}  & 0.319   & 0.310   & 0.298   & 0.301   \\
\hline
\end{tabular}
\end{table}

\begin{table}[!t]
\centering
\caption{Effect of different word embedding initializations for the word-level model: randomly initialized, pre-trained GloVe embeddings on Twitter and Wikipedia. }
\label{word-emb}
\bigskip
\begin{tabular}{l*{4}{c}}
\textbf{Range}  & \multicolumn{2}{c}{\textbf{(0.0-1.0)}} &  \multicolumn{2}{c}{\textbf{(0.5-1.0)}} \\
\textbf{Model}  & \textbf{avg\_p} & \textbf{avg\_s}  & \textbf{avg\_p} & \textbf{avg\_s} \\
\hline
\textbf{Random emb.}       & 0.291    & 0.276    & 0.250   & 0.227 \\
\textbf{GloVe (Twitter)}   & 0.300    & 0.293    & 0.231   & 0.220 \\
\textbf{GloVe (Wiki)}      & \bf{0.326}    & \bf{0.323}    & \bf{0.259}   & \bf{0.252} \\
\hline
\end{tabular}
\end{table}

\subsection{Word-level model}
We used distributed representations to model the words in a tweet. We carried out several experiments where we used random initialization for word embeddings and two pre-trained versions of GloVe embeddings \cite{Pennington2014GloVe:Representation} trained on Wikipedia and Twitter\footnote{\url{https://nlp.stanford.edu/projects/glove/}}, to test if Twitter specific word representations are more suitable to solve the problem. 
Out-of-vocabulary words were replaced with a special word 'OOV' and initialized as a random vector, which was tuned during the training. We used a 50-dimensional embedding representation in all our experiments. 

A bidirectional gated recurrent unit (GRU) network \cite{Chung2014EmpiricalModeling} with a 32-dimension cell size was used for modeling the tweet as a hidden memory vector. The vector corresponding to the last word was fed to a dense layer with 1 neuron predicting emotion intensity. We used GRUs as they tackle the common vanishing gradient problem of RNNs during the training and they contain fewer parameters than LSTM units. 
The word-level model is trained on the given EmoInt corpus with Adam optimizer using different embedding setups, the results are presented in Table \ref{word-emb}.

\begin{table*}[!t]
\centering
\caption{Pearson and Spearman correlation coefficients of baseline, character and word-level models and its ensemble for fear, anger, joy and sadness emotions and also average values. Results are calculated on the provided test set labels.}
\label{all-results}
\begin{tabular}{|c|c|c|c|c|c|c|c|c|c|c|}
\hline
Model                                          & avg\_p & avg\_s & anger\_p & anger\_s & fear\_p & fear\_s & joy\_p & joy\_s & sad\_p & sad\_s \\ \hline
\multicolumn{11}{|l|}{Test set results (Intensity range: 0-1)} \\ \hline
Baseline                                            & 0.655 & 0.652 & 0.631 & 0.623 & 0.631 & 0.622 & 0.645 & 0.654 & 0.712 & 0.711  \\ \hline
\begin{tabular}[c]{@{}c@{}}Char\_LM\end{tabular}    & 0.474 & 0.472 & 0.415 & 0.400 & 0.575 & 0.551 & 0.278 & 0.299 & 0.629 & 0.638  \\ \hline
\begin{tabular}[c]{@{}c@{}}Word\_Level\end{tabular} & 0.326 & 0.323 & 0.253 & 0.258 & 0.337 & 0.332 & 0.201 & 0.194 & 0.435 & 0.395  \\ \hline
\begin{tabular}[c]{@{}c@{}} Char\_LM +\\  Word\_Level\end{tabular} 
                                                    & 0.659 & 0.656 & 0.580 & 0.572 & 0.658 & 0.638 & 0.708 & 0.714 & 0.688 & 0.701   \\ \hline
\begin{tabular}[c]{@{}c@{}}Baseline + \\ Char\_LM +\\  Word\_Level\end{tabular} 
												    & \bf{0.721} & \bf{0.717} & \bf{0.678} & \bf{0.665} & \bf{0.698} & \bf{0.686} & \bf{0.744} & \bf{0.750} & \bf{0.763} & \bf{0.767}   \\ \hline
\multicolumn{11}{|l|}{Test set results  (Intensity range: 0.5-1)} \\ \hline
Baseline                                            & 0.475 & 0.449 & 0.495 & 0.464 & 0.476 & 0.432 & 0.370 & 0.363 & 0.558 & 0.537  \\ \hline
\begin{tabular}[c]{@{}c@{}}Char\_LM\end{tabular}    & 0.419 & 0.413 & 0.316 & 0.327 & 0.488 & 0.435 & 0.416 & 0.423 & 0.457 & 0.467  \\ \hline
\begin{tabular}[c]{@{}c@{}}Word\_Level\end{tabular} & 0.259 & 0.252 & 0.237 & 0.257 & 0.220 & 0.226 & 0.211 & 0.201 & 0.451 & 0.408  \\ \hline
\begin{tabular}[c]{@{}c@{}} Char\_LM +\\  Word\_Level\end{tabular} 
                                                    & 0.471 & 0.467 & 0.389 & 0.406 & 0.488 & 0.435 & 0.536 & 0.547 & 0.470 & 0.481   \\ \hline
\begin{tabular}[c]{@{}c@{}}Baseline + \\ Char\_LM +\\  Word\_Level\end{tabular} 
                                                    & \bf{0.562} & \bf{0.543} & \bf{0.565} & \bf{0.545} & \bf{0.531} & \bf{0.494} & \bf{0.528} & \bf{0.531} & \bf{0.624} & \bf{0.601}   \\ \hline
\end{tabular}
\end{table*}

\section{Experiment}
The dataset for the WASSA-2017 competition \cite{MohammadB17wassa} is comprised of 7097 annotated tweets, classified into 4 categories: joy, anger, fear, and sadness (dataset statistics are presented in Table \ref{dataset-statistics}). For each annotated tweet there is an ID, full text, emotion category, and emotion intensity value. Emotion intensity is a real value in the range from $0$ to $1$, where higher value correspond to a higher intensity of the emotion conveyed.
A sample from the EmoInt corpus:\\
\verb|30112	LOVE LOVE LOVE #smile| 
\verb|#fun #relaxationiskey	joy	0.740|, where 30112 is the ID of a tweet, which is labeled as $"joy"$ with an intensity of 0.740.


\subsection{Ensembling of the models}
Ensembling of several models is a widely used method to improve the performance of the overall system by combining predictions of several classifiers. Several ensembling techniques have been proposed recently: mixing experts \cite{jacobs1991adaptive}, model stacking, bagging and boosting \cite{breiman1996bagging} and a simple weighted average of the scores of individual models, which we used in this work. The main reason for our choice was the limited size of the training data, and using more complex approach like stacking could lead to overfitting.
In this work, we output emotion intensity values as a linear combination of individual predictions of three systems: baseline, character and word-level models.
\begin{multline}
emotion_{intensity} = w_{b}*baseline_{emotion}  \\+ w_{w}*w\_rnn_{emotion} +  w_{c}*c\_rnn_{emotion},\\ w_{b}+w_{w}+w_{c}=1
\end{multline}
where $baseline_{emotion}$, $w\_rnn_{emotion}$ and $c\_rnn_{emotion}$ are intensities of the baseline, character and word-level models correspondingly for the emotion (joy, anger, fear or sadness). 
Ensembling coefficients $w_{b}$, $w_{c}$ and $w_{w}$  were tuned on the development set to maximize the average Pearson correlation coefficient using grid-search. 

\section{Results \& Conclusion}

We report Pearson and Spearman correlation for each emotion class on the provided test data, shown in Table \ref{all-results}. The correlation rank coefficients assess how relevant and similar the two sets of ranking are. The character and word-level neural models achieve lower correlation values than the baseline, which is an indicator that models containing much of external knowledge perform better than end-to-end models on the tasks with a handful amount of samples; however, they bring additional value to the ensemble. Pearson and Spearman correlation coefficients are improved by 0.066 and 0.065 for the intensities in the full range of 0-1, achieving \#4 position on the leaderboard. Additionally, the systems were evaluated on the sample with moderate or high emotional intensities with values from 0.5 to 1. Our ensemble model places rank \#4 and shows 0.087 ($\sim$ 18.5\% relative) improvement on both correlation coefficients.
\par Surprisingly, tweet representations obtained with the character-level model show competitive or even better results for fear and joy emotion categories for samples with high-intensity emotions, and overall the Char\_LM model shows similar results to the AffectiveTweet baseline model. Given the fact that the Char\_LM model did not have any external knowledge or supervision other than the provided data, this demonstrates the effectiveness of the character-level modeling of noisy and short texts.

\section*{Acknowledgments}

This project has received funding from the European Union's Horizon 2020 research and
innovation programme under the Marie Sklodowska-Curie grant agreement No 642667 (SECURE). We would like to thank Dr. Cornelius Weber and Dr. Sven Magg for their helpful comments and suggestions.

\def\urlprefix{}
\def\url#1{}
\bibliography{emnlp2017,Mendeley_yo}
\bibliographystyle{emnlp_natbib}

\end{document}